\documentclass{article}
\usepackage{spconf,amsmath,graphicx}

\usepackage{enumitem}
\usepackage{graphicx}
\usepackage{caption}
\usepackage{subcaption}
\usepackage{tikz}
\usepackage{url}
\usepackage{comment}
\usepackage{indentfirst}
\setlist{nosep, leftmargin=14pt}

\usepackage{mwe} % to get dummy images

\usepackage{hyperref}       % hyperlinks
\usepackage{xcolor}
\usepackage{sidecap}
\hypersetup{
    colorlinks,
    linkcolor={blue!50!black},
    citecolor={blue!50!black},
    urlcolor={blue!50!black}
}

\newcommand{\tikzsubfiglabel}[2][north west]{%
    \node[anchor=#1, inner xsep=2pt, inner ysep=0pt, #2] at (current bounding box.#1) {\footnotesize\strut(\thesubfigure)};%
}
\newcommand{\withsubfiglabel}[2][]{%
    \refstepcounter{subfigure}%
    \begin{tikzpicture}%
        \node[inner sep=0pt, outer sep=0pt] {#2};
        \tikzsubfiglabel{#1}
    \end{tikzpicture}%
}

\title{The Importance of Model Inspection for Better Understanding Performance Characteristics of Graph Neural Networks}
\name{Nairouz Shehata$\,^{1,2}$, Carolina Pi\c{c}arra$\,^1$, Anees Kazi$\,^3$, Ben Glocker$\,^1$}
\address{$^1$ Department of Computing, Imperial College London, UK\\$^2$Biomedical Engineering \& Innovation Laboratory, Aswan Heart Centre, Egypt\\$^3$ Massachusetts General Hospital, Harvard University, USA}
\begin{document}
%\ninept
%
\maketitle
\begin{abstract}
This study highlights the importance of conducting comprehensive model inspection as part of comparative performance analyses. Here, we investigate the effect of modelling choices on the feature learning characteristics of graph neural networks applied to a brain shape classification task. Specifically, we analyse the effect of using parameter-efficient, shared graph convolutional submodels compared to structure-specific, non-shared submodels. Further, we assess the effect of mesh registration as part of the data harmonisation pipeline. We find substantial differences in the feature embeddings at different layers of the models. Our results highlight that test accuracy alone is insufficient to identify important model characteristics such as encoded biases related to data source or potentially non-discriminative features learned in submodels. Our model inspection framework offers a valuable tool for practitioners to better understand performance characteristics of deep learning models in medical imaging.

%This study advances shape classification in neuroimaging by exploring critical architectural decisions in Graph Neural Networks (GNNs), extending our previous work. The methodology incorporates novel graph representations, a multi-graph architecture, and extensive dataset evaluations. The impact of shared and non-shared submodels on graph embedding, the effects of rigid mesh registration as a pre-processing step, and the assessment of multi-graph model generalization capabilities are rigorously examined. Through detailed model dissection, the study uncovers key insights, shedding light on the mechanisms driving discriminating power. These findings offer actionable guidance for practitioners, propelling the application of geometric deep learning in biomedical imaging.

% Our results highlight the significance of model inspection beyond test accuracy, as  as it revealed that overlooking undesirable model characteristics could have detrimental effects on downstream applications.
% Our results show that test accuracy is an insufficient metric for model selection.
\end{abstract}
\begin{keywords}
Shape classification, graph neural networks, brain structures, 3D meshes, model inspection
\end{keywords}
\section{Introduction}
\label{sec:intro}

Understanding biological sex-based differences in brain anatomy provides valuable insights into both neurodevelopmental processes and cognitive functioning. Recent strides in the field of geometric deep learning \cite{bronstein2017geometric}, particularly the advent of Graph Neural Networks (GNNs), have revolutionised the analysis of complex, non-Euclidean data \cite{wu2020comprehensive} to make predictions at a node, edge, or graph-level. This allows us to treat brain shapes as graphs, leveraging the power of GNNs to learn from complex structural anatomical data \cite{sarasua2022hippocampal}. Discriminative feature embeddings can be withdrawn from these models, representing brain shapes as a continuous vector of numerical features that capture valuable structural and geometrical information for downstream prediction tasks \cite{bessadok2022graph}. Techniques like Principal Component Analysis (PCA) can be used to reduce the dimensionality of graph embeddings for visualisation, aiding the exploration of subgroup biases in the feature space beyond the target label. This analysis may help practitioners ensure the reliability of their predictions, and is particularly important in applications where GNNs feature embeddings may be leveraged for new tasks, such as fine-tuning, domain transfer, or multi-modal approaches.

In this study, we dissect GNN models trained under different settings for the task of sex classification using 3D meshes of segmented brain structures. We inspect the learned feature embeddings at different layers within a multi-graph neural network architecture. Through this granular analysis, we reveal critical insights into the inner workings of our models, identifying important effects of different modelling choices. This research demonstrates the utility of conducting a model inspection framework as part of model development, highlighting insights that may guide practitioners in the selection of models with desired characteristics, avoiding biases, overfitting and better understanding the driving forces behind predictions.

%, specially when considering applications such as fine-tuning, domain transfer, or multi-modal approaches, where GCN feature embeddings may be leveraged for new tasks.
%Furthermore, we emphasize the importance of choosing models that prioritize separable feature spaces over those that solely optimize for higher classification accuracy. Such models not only offer improved interpretability, generalization, and robustness but also reduce bias and overfitting, leading to more reliable performance in real-world applications, particularly in sensitive domains like medical imaging.
\section{Methods}
\label{sec:methods}

\subsection{Imaging datasets}
\label{sec:datasets}

We used four neuroimaging datasets, including data from the UK Biobank imaging study (UKBB) \footnote{Accessed under application 12579.} \cite{sudlow2015uk}, the Cambridge Centre for Ageing and Neuroscience study (CamCAN) \cite{taylor2017cambridge, shafto2014cambridge}, the IXI dataset\footnote{{\url{https://brain-development.org/ixi-dataset/}}}, and OASIS3 \cite{lamontagne2019oasis}. Both UKBB and CamCAN brain MRI data were acquired with Siemens 3T scanners. The IXI dataset encompassed data collected from three clinical sites, each employing different scanning systems. CamCAN and IXI are acquired from healthy volunteers, while UKBB is an observational population study. The OASIS3 dataset consists of 716 subjects with normal cognitive function and 318 patients exhibiting varying stages of cognitive decline. For all four datasets, subjects with missing biological sex or age information were excluded. Data from UKBB was split into three sets, with 9,900 scans used for training, 1,099 for validation, and 2,750 for testing. CamCAN, IXI and OASIS3 were used as external test sets, with sample sizes of 652, 563, and 1,034, respectively.

The UKBB data is provided with a comprehensive pre-processing already applied, using FSL FIRST \cite{patenaude2011bayesian} to automatically segment 15 subcortical brain structures from T1-weighted brain MRI, including the brain stem, left/right thalamus, caudate, putamen, pallidum, hippocampus, amygdala, and accumbens-area. We apply our own pre-processing pipeline to the CamCAN, IXI, and OASIS3 datasets, closely resembling the UKBB pre-processing. Our pipeline includes skull stripping using ROBEX \footnote{\url{https://www.nitrc.org/projects/robex}} \cite{iglesias2011robust}, bias field correction using N4ITK \cite{tustison2010n4itk}, and brain segmentation via FSL FIRST.

\subsection{Graph representation}

The anatomical brain structures are represented by meshes as an undirected graph composed of nodes, connected by edges forming triangular faces. The number of nodes for most structures is 642 and up to 1,068, whereas the number of edges per structure ranges between 3,840 and 6,396. The meshes are automatically generated by the FSL FIRST tool.

\subsubsection{Node features}
Each graph node can carry additional information, encoded as feature vectors. This can include spatial coordinates or more complex geometric descriptors. While computer vision has transitioned from hand-crafted features to end-to-end deep learning, we have previously demonstrated the value of using geometric feature descriptors in GNN-based shape classification \cite{shehata2022comparative}. We employ Fast Point Feature Histograms (FPFH) \cite{rusu2009fast}, a pose invariant feature descriptor shown to substantially boost classification performance. To compute FPFH features on a mesh, a point feature histogram is first generated, involving the selection of neighboring points within a defined radius around each query point. The Darboux frame is subsequently defined, and angular variations are computed. This process involves several steps, including the estimation of normals and the calculation of angular variations, resulting in a vector of 33 features at each node.

\subsubsection{Mesh registration}

Mesh registration is an optional pre-processing step, with the goal to remove spatial variability across subjects and datasets. Here, we investigate the use of rigid registration aligning all meshes for a specific brain structure to a standardised orientation using the closed-form Umeyama approach \cite{umeyama1991least}. This method employs a singular value decomposition-based optimisation to obtain an optimal rigid transformation between two given meshes. For each of the 15 brain structures, we select a reference mesh from a random subject from the UKBB dataset, and align the meshes from all other subjects to this reference.  As a result, shape variability due to orientation and position differences is minimised and the remaining variability is expected to primarily represent anatomical differences across subjects.

\begin{figure}[t]
    \centering
    \includegraphics[width=\linewidth]{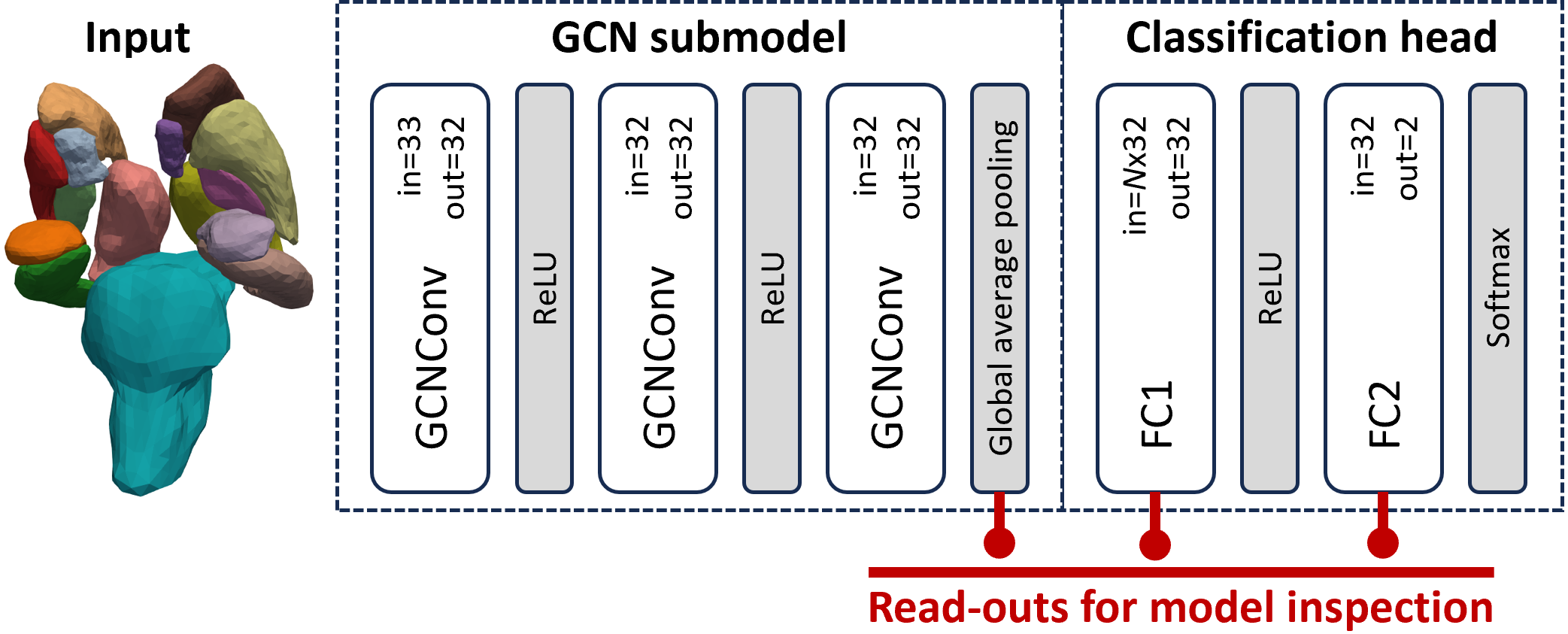}
    \caption{\textbf{Model architecture} consisting of a graph convolutional network (GCN) submodel feeding graph embeddings into a classification head with two fully connected layers (FC1 and FC2). Where N is the number of brain substructures, 15. For our model inspection, we read out the feature vectors from the GCN submodel, FC1, and FC2.}
    \label{fig:model}
\end{figure}

\subsection{Multi-graph neural network architecture}
\label{ssec:Multi-graph Architecture}

Our general GNN architecture is comprised of two main components; the \emph{GCN submodel} which aims to learn graph embeddings over 3D meshes using multiple graph convolutional layers \cite{shehata2022comparative} and an \emph{MLP classification head} that takes the graph embeddings as inputs and performs the final classification using two fully connected layers (cf. Fig. \ref{fig:model}). 

The input to our models are 15 subgraphs representing 15 brain structures, extracted from T1-weighted brain scans. We consider two approaches for learning graph embeddings with GCN submodels. The first approach, referred to as \emph{shared submodel}, uses a single GCN submodel that learns from all 15 subgraphs. Here, the weights of the graph convolutional layers are shared across brain structures. The shared submodel approach is parameter-efficient and aims to learn generic shape features. For the second approach, referred to as \emph{non-shared submodel}, each subgraph is fed into a structure-specific GCN submodel. The non-shared submodel approach has more parameters and may capture structure-specific shape features. In both approaches, the architecture of the GCN submodel is identical and consists of three graph convolutional layers \cite{kipf2016semi} with Rectified Linear Unit (ReLU) activations. A global average pooling layer is used as a readout layer, aggregating node representations into a single graph-level feature embedding. The embeddings from individual structures are stacked to form a subject-level feature embedding which is passed to the classification head.

\subsection{Model inspection}

Our model inspection approach is focused on evaluating the separability of the target label (biological sex, Male and Female) and data source classes (UKBB, CamCAN, IXI or OASIS3) through feature inspection. Each test set sample is passed through the complete pipeline and its feature embeddings are saved at three different stages: at the output layer of the GCN submodel and at the first (FC1) and second (FC2) fully connected layers of the classification head. The dimensions of these embeddings are, respectively,  480 (15 substructures times the hidden layer size, 32), 32 and 2. To allow for visual inspection, the feature embeddings from the GCN and FC1 layers are inputted to a PCA model to reduce their dimensionality. The PCA modes capture the directions of the largest variation in the high-dimensional feature space, allowing us to visualise feature separation in 2D scatter plots. We randomly sample 500 subjects from each dataset for the visualisations. Given that all the models were trained to classify biological sex, a clear separation should be expected between the Male and Female classes in the first PCA modes.

%To conduct a comprehensive model inspection, we evaluate feature separability concerning both the target label (biological sex) and data source. 

%Each sample's feature embedding is saved from the shared/non-shared submodel, resulting in a vector of length 480 (computed as the product of the number of sub-structures, 15, and the hidden size, 32). Similar embeddings are retained for the first and second fully connected layers, denoted as FC1 and FC2, with lengths of 32 and 2, respectively.
%To enhance interpretability, the higher-dimensional features from the GCN and FC1 layers undergo PCA for visualisation in 2D scatter plots. These scatter plots enable a detailed analysis of the effect different modelling choices (shared vs. non-shared submodels, with vs. without mesh registration) on the feature separability.

\section{Experiments \& Results}
\label{sec:results}

%\textbf{Say that we trained and tested four models in total, with and without shared GCN submodels, and with and without mesh registration. Alll models were trained and tested on the same data splits. - ADDED}

For a thorough evaluation, we trained and tested the four models - shared and non-shared GCN submodels, and with and without mesh rigid registration - on identical data splits. All code was developed using PyTorch Geometric and PyTorch Lightning for model implementation and data handling. We used the Adam optimiser \cite{kingma2014adam} with a learning rate of 0.001, and employed the standard cross entropy loss for classification. Random node translation was used as a data augmentation strategy with a maximum offset of 0.1mm \cite{zhou2020data}. This was shown to improve performance in our previous study \cite{shehata2022comparative}. Model selection was done based on the loss of the validation set. Our code is made publicly available\footnote{\url{https://github.com/biomedia-mira/medmesh}}.

\subsection{Classification performance}
\label{ssec:generalization}

%\textbf{We first look at the overall classification performance. See Fig. \ref{fig:model-performance}. We can see some effect of using mesh registration. Closing the gap between in-distribution UKBB test data and the external test data (CamCAN, IXI, OASIS3). Very little difference between shared vs non-shared submodels. Generally, very little differences, and given that model selection might be done by looking at in-distribution test data (UKBB), a practitioner might opt for model `shared without registration' for convenience and smaller number of parameters. But, as we will see, test accuracy is insufficient to highlight differences in the model characteristics.- ADDED}

\begin{figure*}[t]
    \centering
    \begin{subfigure}{0.24\linewidth}
        \centering
        \withsubfiglabel[xshift=-5pt]{
            \includegraphics[width=\linewidth]{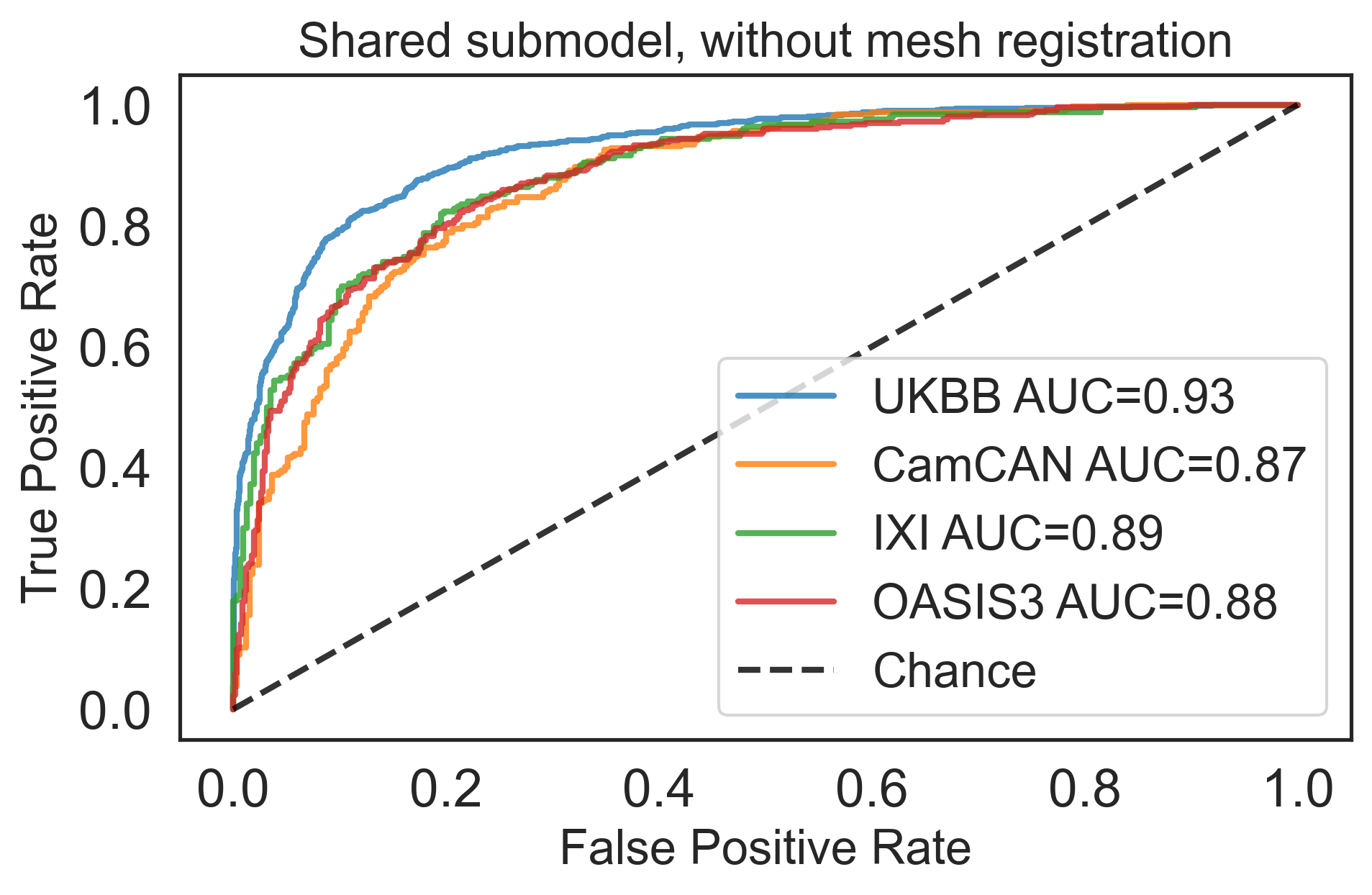}}
    \end{subfigure}
    \begin{subfigure}{0.24\linewidth}
        \centering
        \withsubfiglabel[xshift=-5pt]{
            \includegraphics[width=\linewidth]{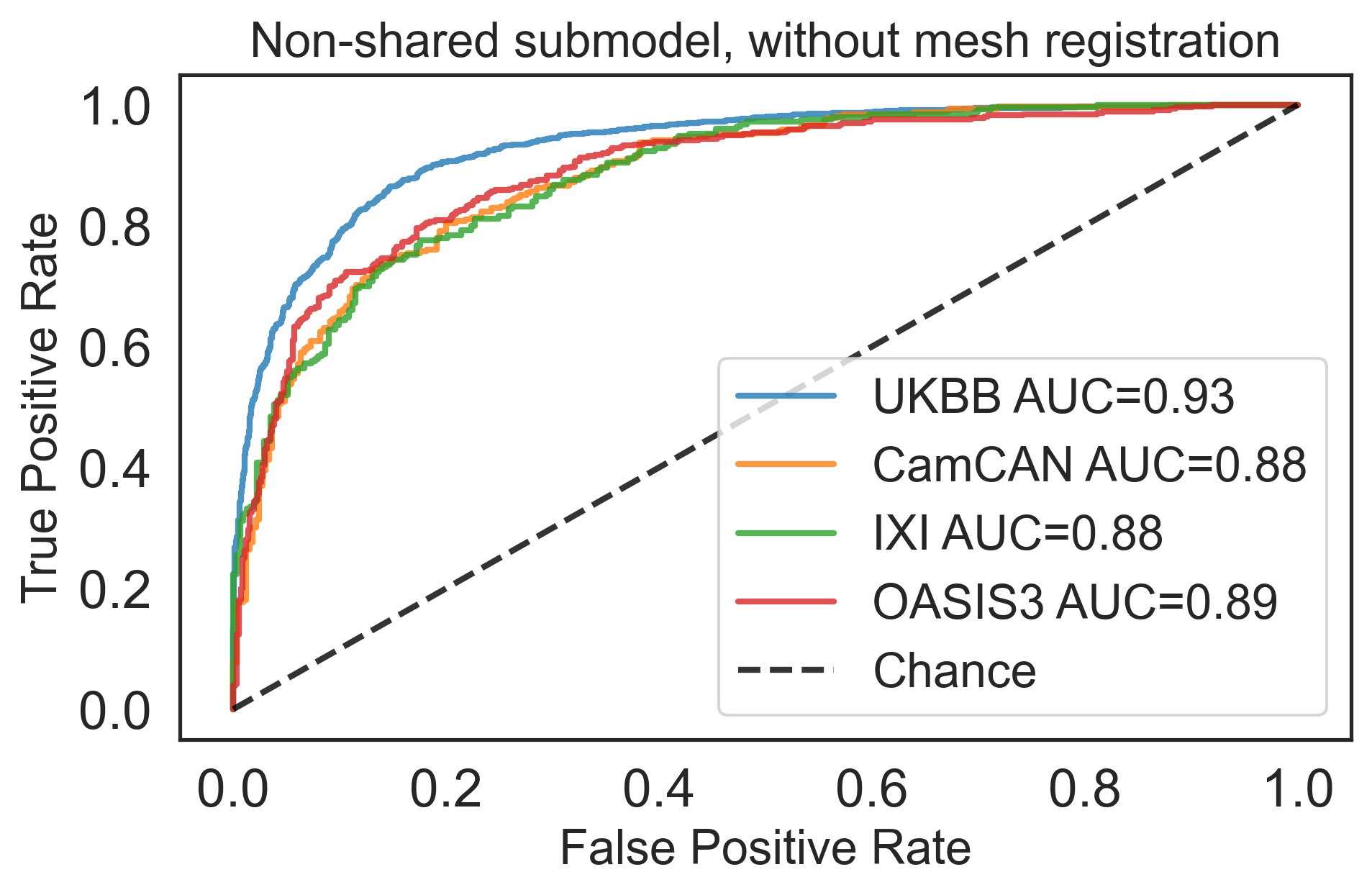}}
    \end{subfigure}
    \begin{subfigure}{0.24\linewidth}
        \centering
        \withsubfiglabel[xshift=-5pt]{
            \includegraphics[width=\linewidth]{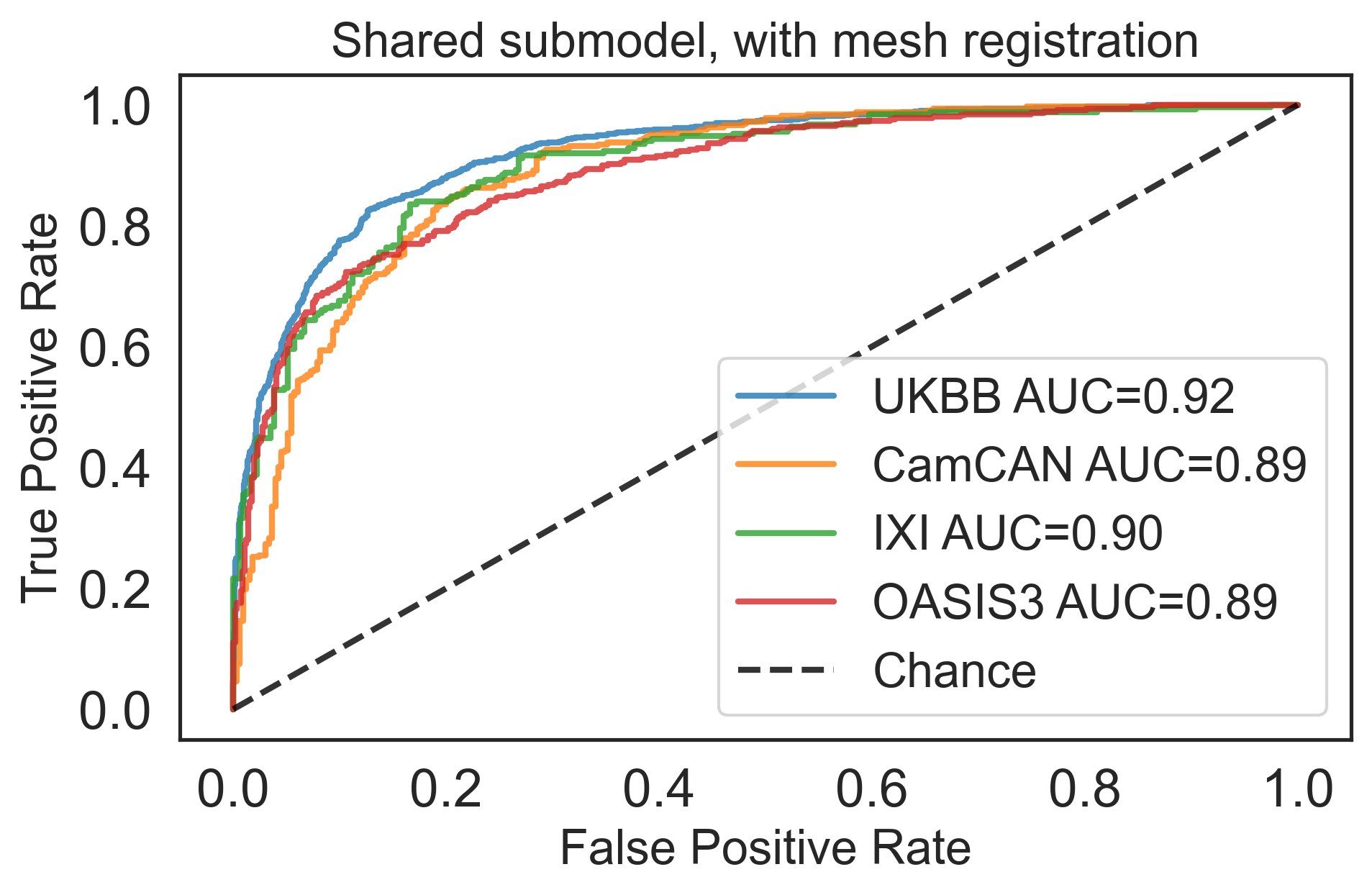}}
    \end{subfigure}
    \begin{subfigure}{0.24\linewidth}
        \centering
        \withsubfiglabel[xshift=-5pt]{
            \includegraphics[width=\linewidth]{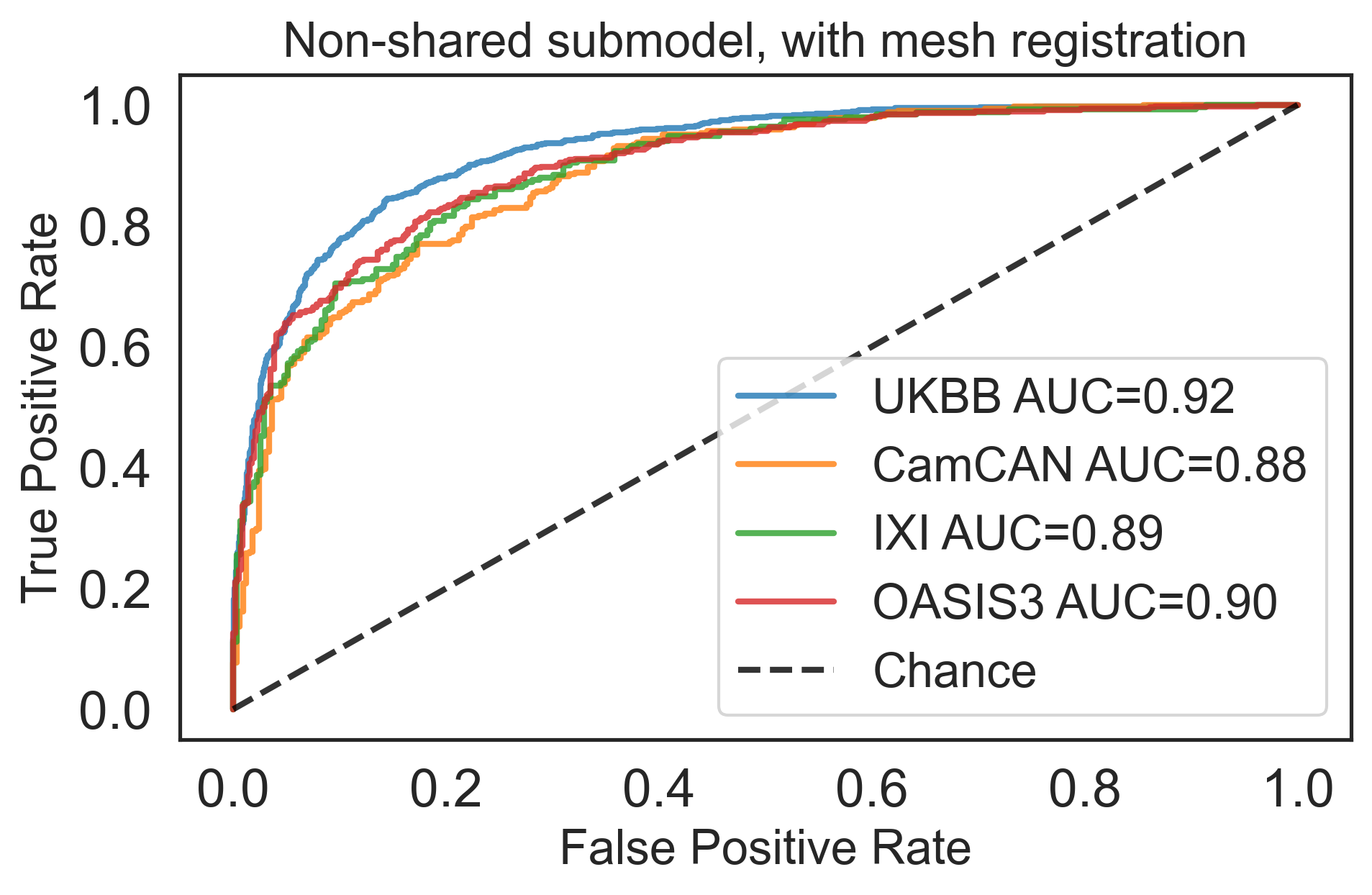}}
    \end{subfigure}
    \caption{\textbf{Sex classification performance} for four models; \textbf{(a)} \emph{shared} and \textbf{(b)} \emph{non-shared} submodel \emph{without} mesh registration, \textbf{(c)} \emph{shared} and \textbf{(d)} \emph{non-shared} submodel \emph{with} mesh registration. We observe that the generalisation gap between the in-distribution test data (UKBB) and the external test data (CamCAN, IXI, OASIS3) closes \emph{with} mesh registration. Overall, there are only small differences in performance, illustrating that test accuracy alone is insufficient to identify variations in model characteristics.}
    \label{fig:model-performance}
\end{figure*}

Figure \ref{fig:model-performance} summarises the classification performance of the four models, showing the ROC curves together with the area under the curve (AUC) metric, reported separately for each of the four test datasets. There are two main observations: (i) There are very little differences in the absolute performance across the four models. Comparing the \emph{shared} vs \emph{non-shared} submodel, the AUC performance is comparable. When comparing models \emph{with} and \emph{without} mesh registration, we find the generalisation gap decreases between in-distribution test (UKBB) and the external test data (CamCAN, IXI, OASIS3). However, we also observe a small drop in performance on the in-distribution test data when using mesh registration, compared to not using registration. A practitioner using internal test results for final model selection may opt for using a \emph{shared} submodel, due to its parameter efficiency, \emph{without} mesh registration, due to convenience. As we will see next, this choice may be suboptimal as test accuracy alone is insufficient to identify important model characteristics.
%The ROC curves in Fig. \ref{fig:model-performance} depict the classification performance of each of the four models on both the in-distribution UKBB test set and external test data (CamCAN, IXI, and OASIS3). We observe that applying rigid registration as a data preprocessing step closes the generalisation gap between the UKBB test data and the external test data.
%Overall, there are only small differences in performance, when measured by test accuracy,illustrating it is insufficient to identify variations in model characteristics. 

%Choosing a model based solely on UKBB test accuracy might favor discarding mesh registration, and choosing a shared submodel given its fewer training parameters. However, this choice will be challenged via the scatter plots from the proposed model inspection approach in the subsequent subsections.

%When using either the shared or non-shared submodel, we observe the model effectively separating the datasets, particularly evident in Figure \ref{fig:res3} a and b for the UKBB dataset. However, rigid registration of the meshes seems to mitigate this dataset/domain separability, as can be observed in Figure X. 

% upon applying rigid registration of the meshes prior to training, we note that the datasets no longer exhibit apparent separability. Rigid-registration appears to have mitigated the datasets’ differences.

\subsection{Effect of using structure-specific submodels}
\label{sec:submodel}

For the models that use a \emph{shared} submodel, we observe that the GCN feature embeddings are non-discriminative with respect to the target label. Separation seems completely missing in the \emph{shared} model \emph{without} registration (see Fig. \ref{fig:feature-inspection}a), with only weak separation in the \emph{shared} model \emph{with} registration (see Fig. \ref{fig:feature-inspection}c). For these models, the classification heads will primarily contribute to the model performance. For the models with a \emph{non-shared} submodel, we find a much better separability for the GCN features \emph{with} and \emph{without} mesh registration (cf. Figs. \ref{fig:feature-inspection}b, d). Here, the GCN features will meaningfully contribute to the models' classification performance.

\subsection{Effect of mesh registration}
\label{sec:registration}

%\textbf{Let's first discuss the effect of mesh registration. We can clearly see that without mesh registration, the GCN feature embeddings from the submodel strongly encode data source (separate clusters for UKBB and external test data), instead of the target label. Compare Figs. \ref{fig:feature-inspection-sex} and \ref{fig:feature-inspection-dataset}. This is observed both for the shared and non-shared submodels. Dataset is more strongly encoded that the actual target label. The classification performance of the models without registration seems to primarily stem from the classification head, as there is clear separation in FC1 and FC2 feature vectors.- ADDED}

\begin{figure*}[!h]
    \centering
    \vspace{5mm}
    \begin{subfigure}{0.48\linewidth}
        \centering
        {
            \includegraphics[width=\linewidth]{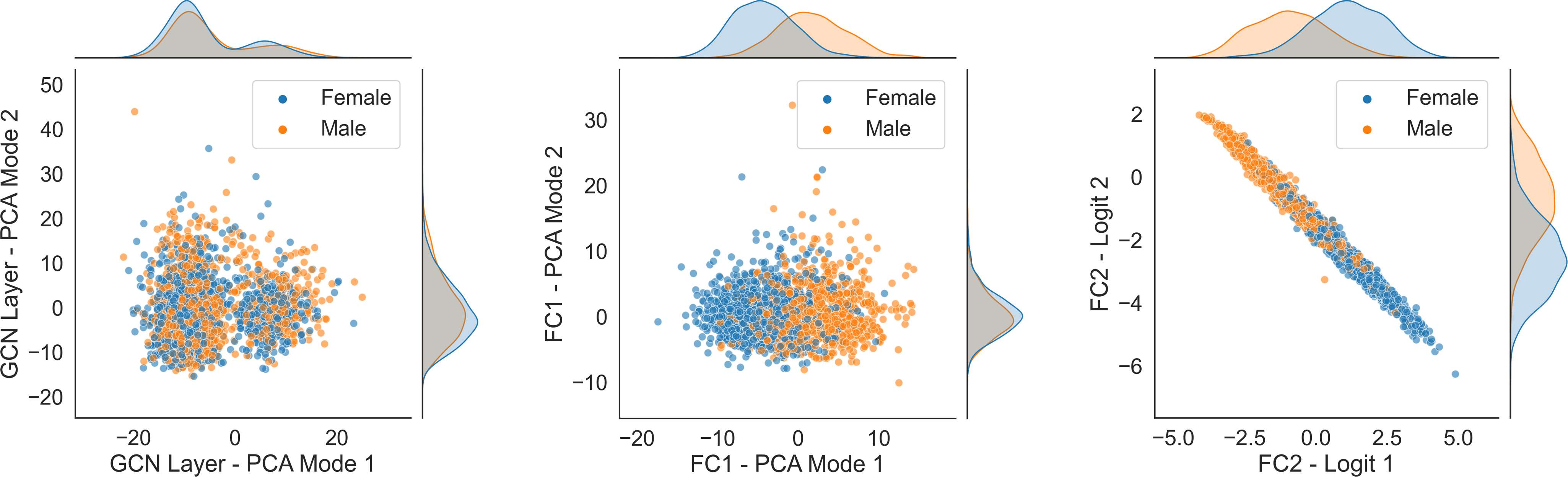}}
    \end{subfigure}
    \hspace{4mm}
    \begin{subfigure}{0.48\linewidth}
        \centering
        {
            \includegraphics[width=\linewidth]{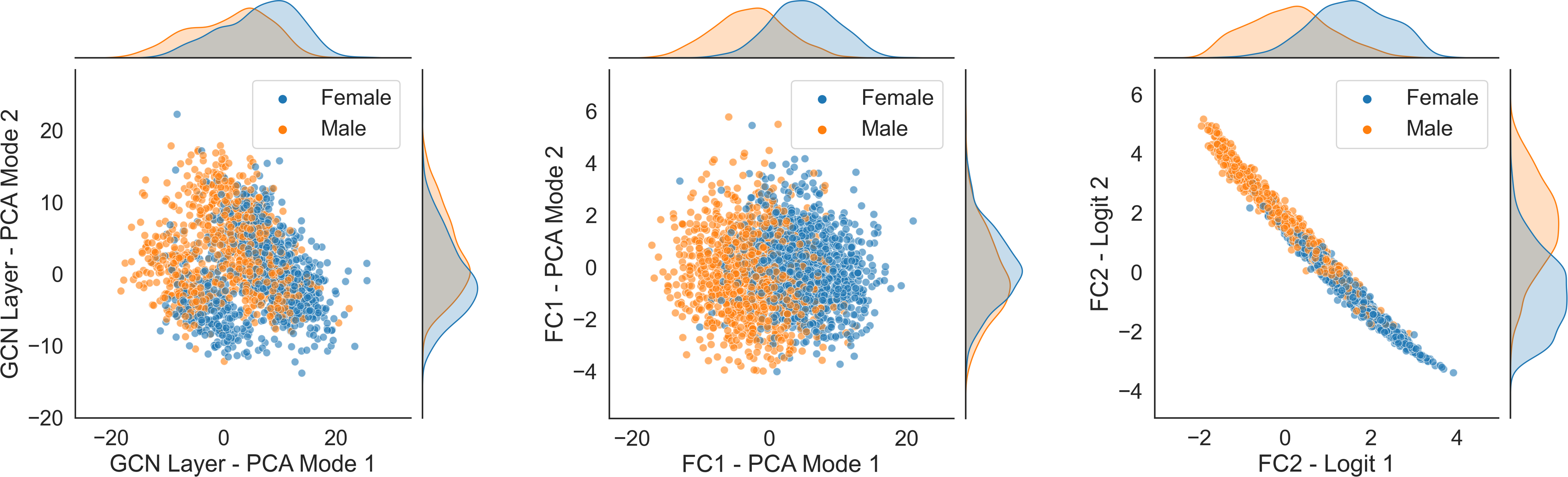}}
    \end{subfigure}
    \begin{subfigure}{0.48\linewidth}
        \centering
        \vspace{2mm}
        {
            \includegraphics[width=\linewidth]{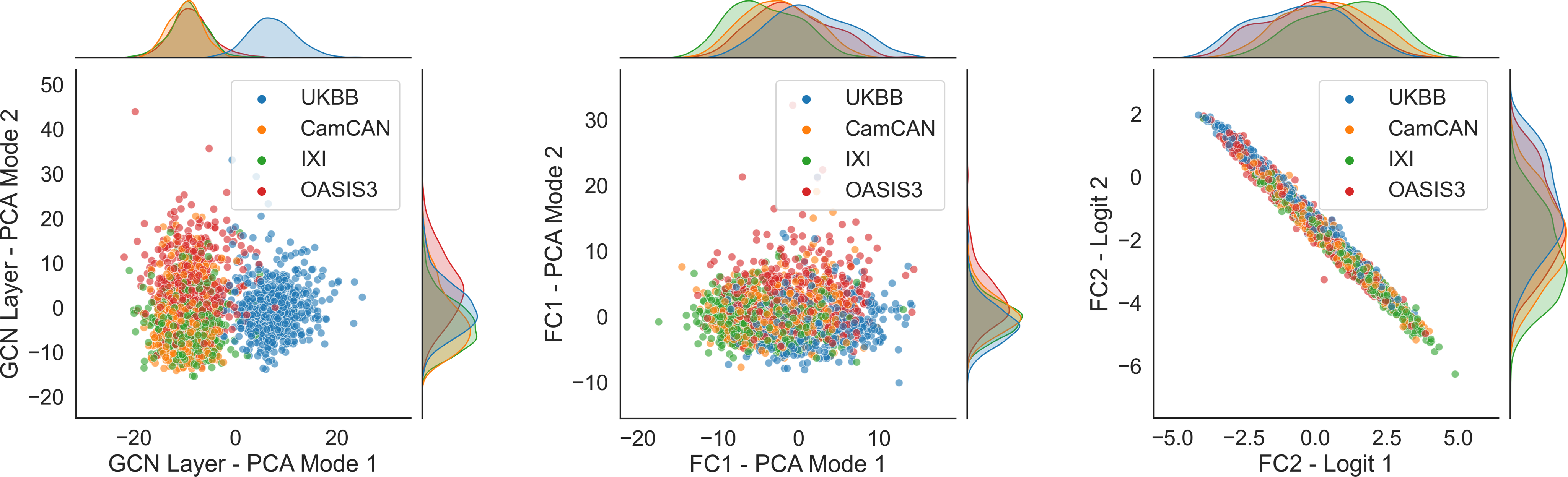}}
        \caption{\emph{Shared} submodel, \emph{without} mesh registration}
    \end{subfigure}
    \hspace{4mm}
    \begin{subfigure}{0.48\linewidth}
        \centering
        {
            \includegraphics[width=\linewidth]{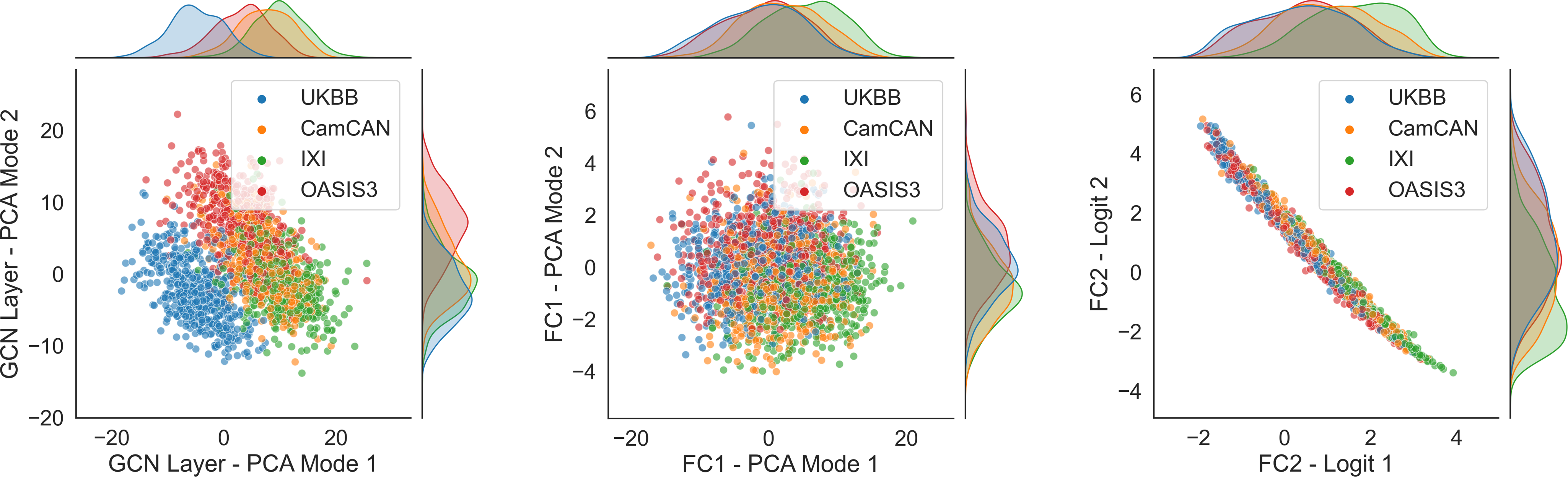}}
        \caption{\emph{Non-shared} submodel, \emph{without} mesh registration}
    \end{subfigure}
    \begin{subfigure}{0.48\linewidth}
        \vspace{6mm}
        \centering
        {
            \includegraphics[width=\linewidth]{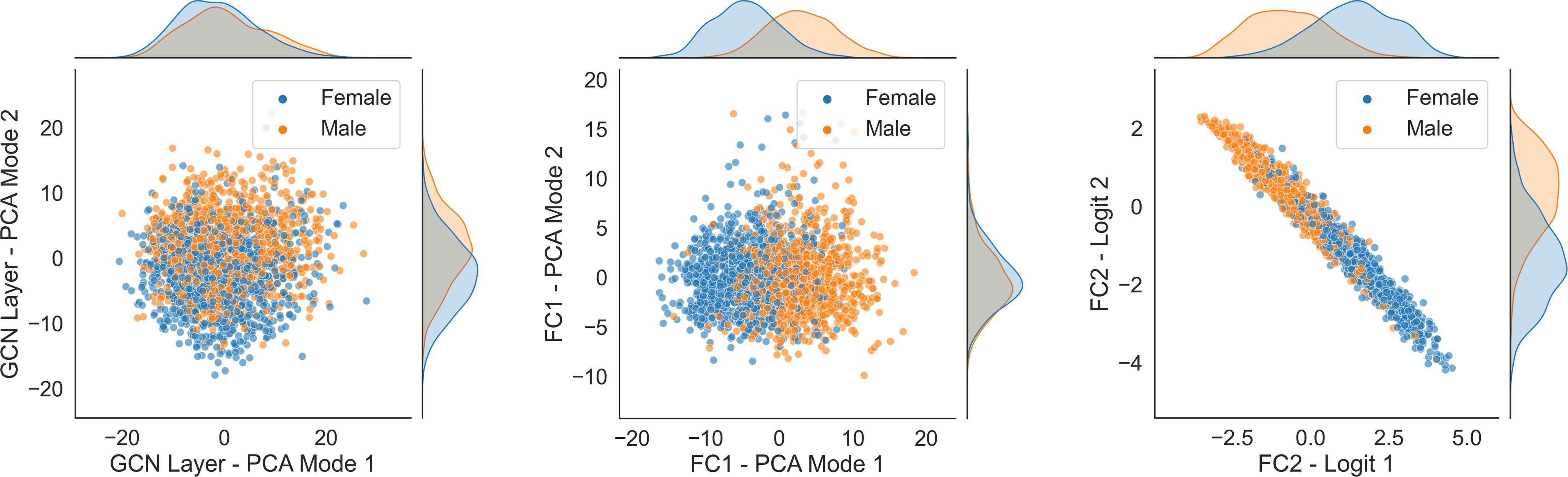}}
    \end{subfigure}
    \hspace{4mm}
    \begin{subfigure}{0.48\linewidth}
        \centering
        {
            \includegraphics[width=\linewidth]{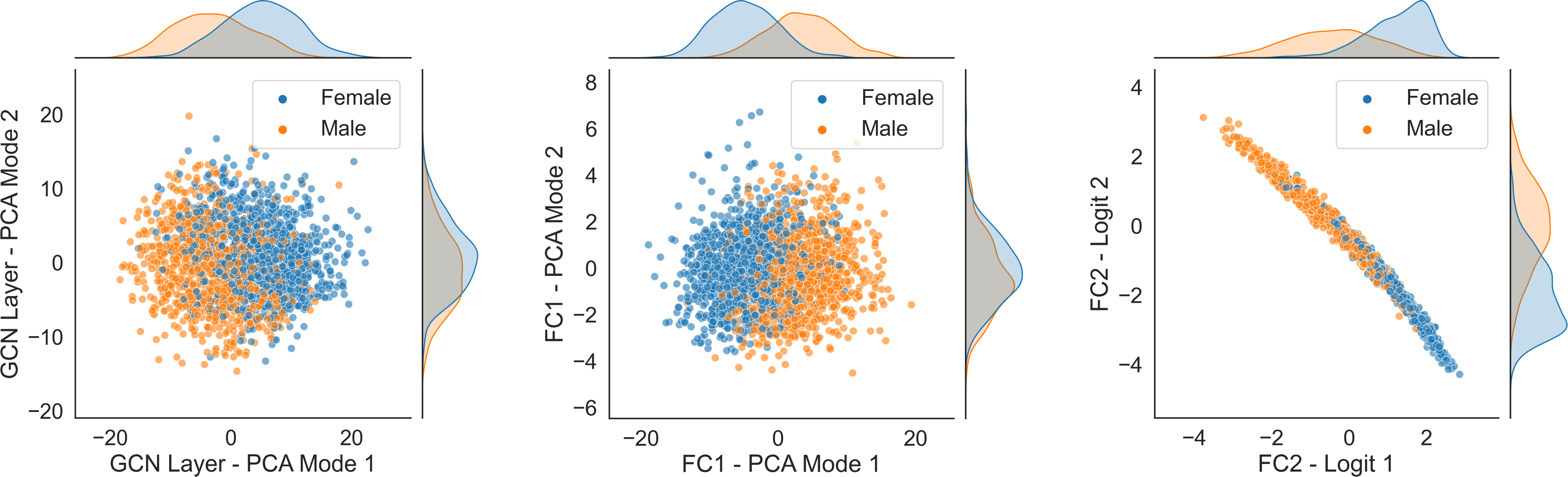}}
    \end{subfigure}
    \begin{subfigure}{0.48\linewidth}
        \vspace{2mm}
        \centering
        {
            \includegraphics[width=\linewidth]{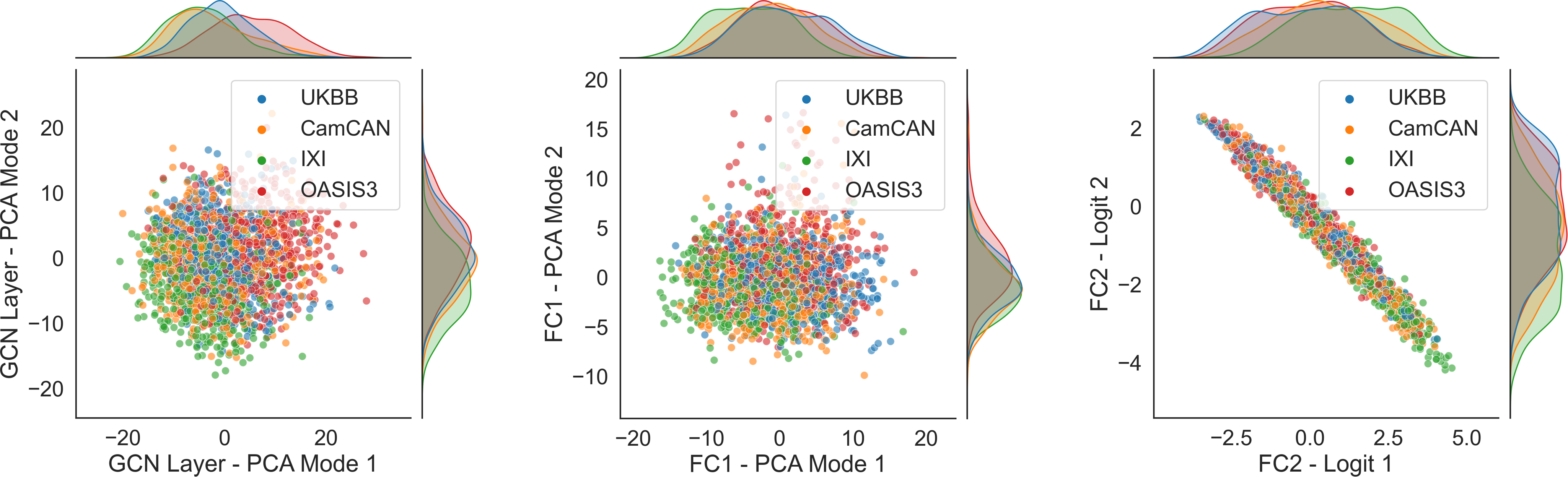}}
        \caption{\emph{Shared} submodel, \emph{with} mesh registration}
    \end{subfigure}
    \hspace{4mm}
    \begin{subfigure}{0.48\linewidth}
        \centering
        {
            \includegraphics[width=\linewidth]{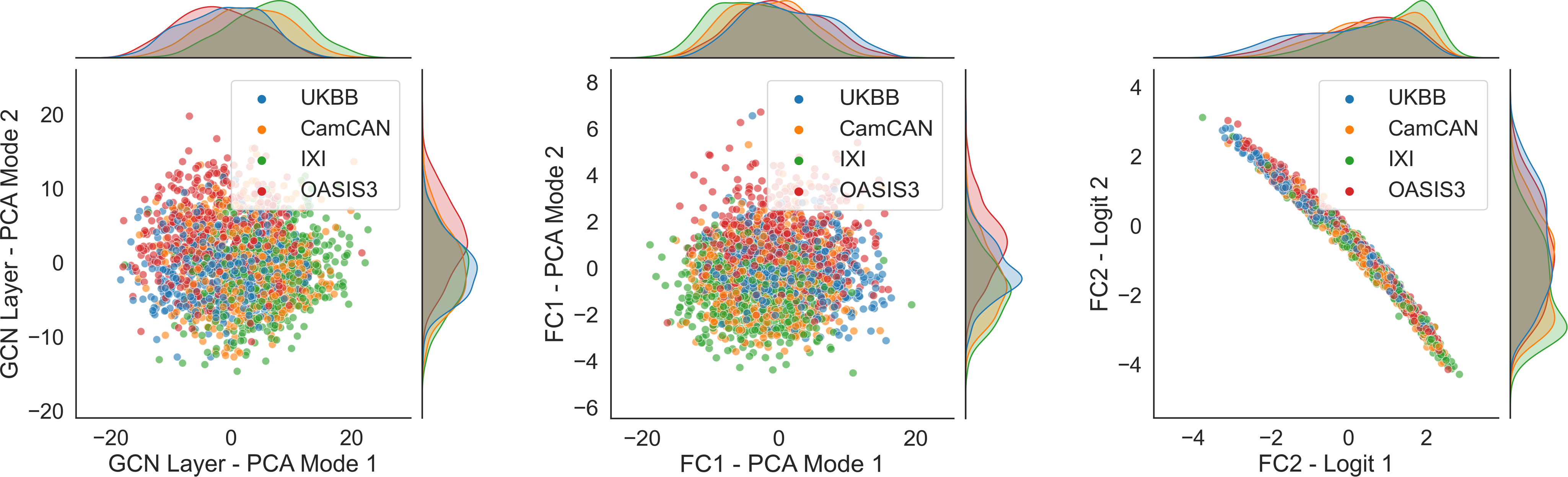}}
        \caption{\emph{Non-shared} submodel, \emph{with} mesh registration}
    \end{subfigure}
    \caption{\textbf{Effect of modelling choices on feature separability} for four different models at their the GCN layer (left), first fully connected layer FC1 (middle), and output layer FC2 (right). \textbf{Models}: \textbf{(a,c)} \emph{shared} and \textbf{(b,d)} \emph{non-shared} GCN submodel, and \textbf{(a,b)} \emph{without} and \textbf{(c,d)} \emph{with} mesh registration. For each model, we show the separation by target label in the top row, and the separation by dataset in the bottom row. \textbf{Effect of submodel:} The models in \textbf{(a,c)} with a \emph{shared} submodel are unable to learn discriminative GCN features for the prediction task, while the models in \textbf{(b,d)} with a \emph{non-shared} submodel show much better task-related separability in the GCN features. \textbf{Effect of registration:} The models models in \textbf{(a,b)} \emph{without} registration strongly encode information about the data source in the GCN layer. This is much reduced for the models in \textbf{(c,d)} \emph{with} mesh registration.}
    \label{fig:feature-inspection}
\end{figure*}

When studying the effect of mesh registration, we can clearly observe that \emph{without} registration, the GCN feature embeddings from the submodel strongly encode data source, showing separate clusters for UKBB and external test data (cf. Figs. \ref{fig:feature-inspection}a,b). When introducing mesh registration as a pre-processing step, we note a significant improvement, with an almost entirely removed separation of datasets in the GCN layer independent of whether a \emph{shared} and \emph{non-shared} submodel is used (Figs. \ref{fig:feature-inspection}c, d). The separability of the target label in the GCN layer is well defined for the \emph{non-shared} submodel (Fig. \ref{fig:feature-inspection}d), while remaining weak for the \emph{shared} submodel (Fig. \ref{fig:feature-inspection}c). Rigid registration as a pre-processing step seems to not only improve the learning efficiency of the GCN submodel, but also its ability to generalise across data distributions.

\section{Conclusion}
\label{sec:conclusion}

Our findings underscore the limitations of relying solely on test accuracy for model selection, particularly when focusing on in-distribution test accuracy. We demonstrate that this may lead practitioners to select models with undesired characteristics where GCN features are non-discriminative for the prediction task and/or strongly encode biases such as data source. Using a comprehensive model inspection, we were able to identify variations in the model characteristics and better understand what drives the final prediction (GCN submodel vs classification head). The importance of this becomes evident when considering applications such as fine-tuning, domain transfer, or multi-modal approaches, where GCN feature embeddings may be leveraged for new tasks. %In this particular case, the use of the feature embeddings from the model with shared GCN submodels would have either a negligible or harmful effect on a downstream application.

%Our results show how model design, data preparation, and classification interact in a complex manner, emphasizing the utility and value of model inspection. Here, we used the example task of sex classification, as it allowed us to study performance differences across different datasets. 
Our model inspection framework can be easily applied to other models, tasks, and purposes. It was previously used to detect biases in chest radiography disease detection models \cite{glocker2023risk}. Here, we strongly advocate for the wider use of model inspection as an integral part of comparative performance analyses. We hope that our work can contribute to improving the reliability of model selection in all areas of deep learning for biomedical image analysis.

%The shared submodel shows similar performance to non-shared, while the non-shared submodel displays clearer separability in the embeddings. Rigid mesh registration reduces distinctiveness between datasets. Most of the learning took place in the fully connected layers only when using a shared submodel.

%These results emphasise how model design, data preparation, and classification interact in a complex manner. Our study offers crucial insights for practitioners, providing actionable guidance for optimising GNN-based shape classification in neuroimaging.

\section{Acknowledgments}
\label{sec:acknowledgments}

Nairouz Shehata is grateful for the support by the Magdi Yacoub Heart Foundation and Al Alfi Foundation.

\section{Compliance with ethical standards}
\label{sec:ethics}

This study uses secondary, fully anonymised data which is publicly available and is exempt from ethical approval.

\bibliographystyle{IEEEbib}
\bibliography{strings,refs}

\end{document}